\documentclass[sn-mathphys,iicol]{sn-jnl}

\jyear{2021}%

\theoremstyle{thmstyleone}%
\theoremstyle{thmstyletwo}%

\theoremstyle{thmstylethree}%

\usepackage[dvipsnames]{xcolor}
\colorlet{revision}{black} 
\usepackage{tabularx}
\usepackage{tablefootnote}

\begin{document}

\title[Z. Liu et al]{Curiosity-Diffuser: Curiosity Guide Diffusion Models for Reliability}


\author[1]{\fnm{Zihao} \sur{Liu}}

\author[1]{\fnm{Xing} \sur{Liu}}

\author[2]{\fnm{Yuhang} \sur{Dong}}
\author[1]{\fnm{Haitao} \sur{Chang}}

\author[1]{\fnm{Zhengxiong} \sur{Liu}}

\author[1]{\fnm{Panfeng} \sur{Huang}}

\affil[1]{\orgdiv{School of Astronautics}, \orgname{Northwestern Polytechnic University}, \orgaddress{\city{Shaanxi} \postcode{710072},  \country{China}}}

\affil[2]{\orgdiv{Polytechnic Institute}, \orgname{Zhejiang University}, \orgaddress{\city{Zhejiang} \postcode{310027},  \country{China}}}


\abstract{One of the bottlenecks in robotic intelligence is the instability of neural network models. This leads to risks when applying intelligence in the physical world. Specifically, imitation policy based on neural network may generate hallucinations, leading to inaccurate behaviors that impact the safety of real-world applications. 
To address this issue, this paper proposes the Curiosity-Diffuser, aimed at guiding the conditional diffusion model to generate trajectories with lower curiosity, thereby improving the reliability of policy. The core idea is to use a Random Network Distillation (RND) curiosity module to assess whether the model's behavior aligns with the training data, and then minimize curiosity by classifier guidance diffusion to reduce overgeneralization during inference. 
Additionally, we propose a computationally efficient metric for evaluating the reliability of the policy, measuring the similarity between the generated behaviors and the training dataset, to facilitate research about reliability learning.
Finally, simulations and real-world experiments verify the effectiveness and applicability of the proposed method to a variety of scenarios, showing that Curiosity-Diffuser significantly improves task performance and produces behaviors that are more similar to the training data. The code for this work is available
at: \href{github.com/CarlDegio/Curiosity-Diffuser}{github.com/CarlDegio/Curiosity-Diffuser}}




\maketitle

\section{Introduction}\label{sec1}

Recent years have witnessed remarkable progress in the field of robotic imitation learning, driven by breakthroughs in deep neural networks and large-scale demonstration datasets.
Early approaches focusing on direct Behavior Cloning(BC) have evolved into sophisticated frameworks capable of learning from multimodal perception and multimodal action. Notable examples include transformer-based methods \cite{aloha}, diffusion policies based on diffusion models \cite{diffusion_policy}, and Vision-Language-Action(VLA) methods employing large neural networks \cite{3d_vla}. Their appeal stems from the ability to easily scale across different robot architectures for diverse tasks through training with newly collected task data, without requiring significant algorithmic modifications.
These advancements have facilitated a range of real-world applications, such as open-world quadruped robots capable of throwing balls \cite{umi_leg}, underwater robots engaged in debris collection \cite{water_robot}, and humanoid robots executing dance movements \cite{asap_dancing}.

However, these methods remain imperfect, as imitation learning systems often exhibit limited reliability when faced with complex tasks or multi-task scenarios. Empirical studies reveal notable performance gaps. For instance, the action chunking transformer framework achieves success rates of approximately 80\% in both simulation tasks and real-world deployments when trained with human data \cite{aloha}. Similarly, the OpenVLA model \cite{openvla}, despite leveraging massive training datasets, demonstrates an average task success rate of merely 70\%, while the multi-task 3D Diffusion Policy \cite{3d_diffusion_policy} achieves success rates exceeding 50\% in only 20\% of its designated tasks. 
{\color{revision}Moreover, such instability can be amplified during closed-loop execution. In the context of imitation learning, we use the term hallucination to refer to unreliable action or trajectory generation when a policy encounters out-of-distribution states that are not sufficiently covered by the training dataset. Unlike an isolated prediction error, such behavior can be recursively amplified in robot control: an inaccurate action changes the next state, and the resulting state mismatch may further move the policy away from the training distribution, leading to compounding errors \cite{compounding_error, aloha}.}
{\color{revision}This reliability deficit poses a critical barrier to real-world robot adoption, as potential failures in mission-critical applications raise fundamental safety concerns.}

To address these reliability challenges, various technological approaches have been explored in the research community. For instance, works such as RT-X \cite{RT-X} and RDT \cite{rdt} aim to establish larger datasets to overcome the long-tail distribution issue and generate scaling laws similar to those observed in large language models. The HIL-SERL \cite{HIL_SERL} work introduces reward values through reinforcement learning fine-tuning for imitation tasks to provide preferences for optimal strategies, thereby increasing the data density near the optimal strategy and enhancing the performance of neural networks. Meanwhile, innovative applications of curiosity-driven methods have been observed in the field of reinforcement learning. Curiosity methods evaluate the novelty of data relative to the model, guiding the exploration process to produce additional new data. \textcolor{revision}{In contrast, for imitation learning, novelty estimation can be used in the opposite direction: discouraging trajectories that move into out-of-distribution regions during decision-making, thereby reducing unreliable generation and compounding errors. Essentially, this can be regarded as using a curiosity model to evaluate the reliability of the behavior generated by the imitation learning model, representing a form of model evaluation.}


\begin{figure}[thpb]
\centering
\includegraphics[width=0.45\textwidth]{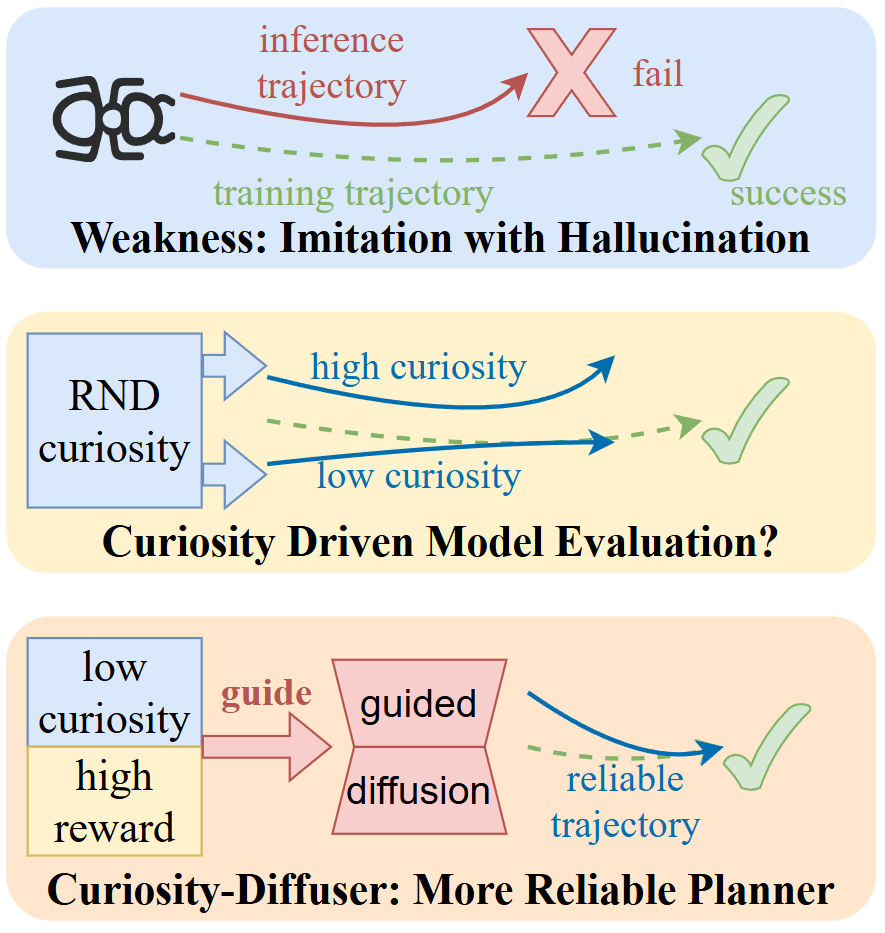}
\caption{
The motivation behind the Curiosity-Diffuser method is that imitation learning can produce hallucinations due to model inference instability when facing new states. We observed that curiosity mechanisms can evaluate the novelty of states. Therefore, based on the guided diffusion framework, we propose the Curiosity-Diffuser method, which combines curiosity-based guidance and reward-based guidance. Through experiments, we demonstrate that the introduction of the curiosity mechanism enhances the reliability of execution.
}
\label{brief}
\end{figure}

In this paper, we propose a general method, termed Curiosity-Diffuser, to enhance the safety and performance of imitation learning. 
The core idea, as illustrated in Fig. \ref{brief}, is to train a curiosity module based on Random Network Distillation(RND) \cite{rnd} that guides the conditional diffusion model to generate trajectories exhibiting lower curiosity, thereby enabling more reliable decision-making. Specifically, we posit that the policy model can achieve a good fit on the training data, making it crucial to avoid \textcolor{revision}{compounding error induced drift into out-of-distribution regions} during inference. To this end, we draw inspiration from curiosity mechanisms in reinforcement learning and use them to assess whether the policy output has been previously observed in the training dataset. Furthermore, since the conditional diffusion model policy is capable of planning based on specific preferences \cite{diffuser}, we utilize its classifier guidance to implement curiosity-minimized planning, with the training of the policy network and the classifier network decoupled. Finally, to verify our approach’s effectiveness, we introduce an indicator that evaluates the reliability of imitation learning test results by measuring their closeness to the dataset, thereby promoting research on the stable execution of imitation learning.

The contributions of our work are as follows:
\begin{itemize}
\item We propose a method for evaluating model reliability based on the principle of curiosity.
\item Leveraging the conditional diffusion model, we develop a curiosity-guided diffusion approach for reducing hallucinations of the imitation learning.
\item We introduce a novel indicator, K-Sim, to effectively measure whether the decisions are sufficiently similar to the training dataset to avoid hallucinations.
\item We conduct extensive simulations and real-world experiments to evaluate the effectiveness of our proposed method in generating more reliable behaviors while also enhancing task performance.
\end{itemize}

\section{Related Works}

\subsection{Imitation Learning}

Imitation learning has recently led to many impressive works, especially in the field of robotics, where there has been a surge of research focusing on high-performance imitation learning using large datasets and models \cite{imitation_survey}. For example, the ManipLLM \cite{manipllm} work fine-tunes the LLAMA model, leveraging the knowledge from pre-trained models to enable imitation learning to perform well on unseen tasks. The $ \pi_0 $ method \cite{pi_0} is fine-tuned on a visual language model (VLM), resulting in a relatively general robot policy that is tested on a large number of new tasks in a zero-shot manner. However, such methods are still somewhat limited by the coverage of the dataset and lack task execution preferences, which means they cannot surpass the policy performance at the time of data collection.


Reinforcement learning (RL) introduces a new perspective to imitation learning by providing human preferences for the environment in the form of reward values, compensating for the data scarcity caused by imitation learning's reliance solely on positive samples. In particular, offline reinforcement learning (Offline RL) \cite{offline_rl} enables strategy learning using pre-collected trajectory data and rewards, making it possible to form a hybrid policy by stitching together the good parts of trajectories based on rewards, as done in the decision transformer \cite{decision_transformer}. In particular, recent work such as diffusion Q-learning \cite{diffusion_qlearning} attempts to combine diffusion models with Q-learning, showcasing a stronger representation capability. However, offline reinforcement learning still faces the issue of value overestimation due to the distribution mismatch between the offline dataset and the learned strategy, which makes methods with constraints \cite{offlinerl_implicit_regular} crucial to ensure model stability in offline RL.


Compared to offline RL, we focus more on imitation learning because it does not require designing complex rewards dependent on states, thus enhancing its potential for practical application. However, we can introduce simple preferences into imitation learning based on the success of trajectories and add evaluation modules to ensure the reliability of its reasoning.


\subsection{Neural Network Evaluation}

\textcolor{revision}{The inherent instability of neural networks as open-loop executors has been extensively studied across many application domains \cite{stable_bc, aloha, BC_auto_drive}. In natural language processing, this instability manifests as hallucination phenomena in large language models (LLMs), which lead to factually inconsistent outputs \cite{hallucination_survey}. Eliminating such hallucinations is crucial for improving the performance of policies.} Recent research has focused on the activation patterns of neural networks, using two metrics, Logit Lens and Tuned Lens, for each layer of the network to assess whether the model's current output is reliable \cite{llm_qbq}. Approaching this from the perspective of entropy can also yield effective results. For instance, entropy can be used to statistically analyze the activation patterns of the neural network across the entire dataset to gauge whether the network has learned the complex patterns within the data \cite{nips_entropy}. Furthermore, entropy can be used to detect the cognitive uncertainty of the model, as higher cognitive uncertainty increases the likelihood of the model generating hallucinations \cite{dm_entropy}.

In robotic imitation learning, similar instability can arise when the policy encounters unfamiliar states outside the training distribution, leading to compounding errors \cite{compounding_error}. However, unlike NLP problems where standard answers are available, it is challenging to estimate whether the output is reliable in these situations. The conformal prediction method \cite{conformal_prediction} addresses this by introducing a calibration set with a specified confidence level. By comparing the classification probabilities or error magnitudes between the calibration set and the test set, it ensures that the output during testing meets the reliability requirements.
We noticed that curiosity-driven methods used to enhance exploration in reinforcement learning, such as RND \cite{rnd} and Intrinsic Curiosity Module (ICM) \cite{ICM} introduce additional terms that measure the novelty of states. These terms act as extra reward signals to encourage agents to explore, thereby enabling reinforcement learning in sparse reward tasks. This functionality is precisely suited to evaluate the reliability of the model's output. Therefore, the focus of this work is to utilize a simple RND approach to estimate the confidence of neural networks by fitting random priors. This method has been proven to be effective and superior to model ensembling \cite{rnd_better_ensemble}. 

\subsection{Guided Diffusion}
The diffusion model is a powerful generative model that represents the data generation process as an iterative denoising process \cite{diffusion_model, diffusion_tutorial, dpm-solverpp}. It has been widely used in image generation \cite{stable_diffusion, diffusion_synthesis_survey} and robot decision making \cite{diffusion_policy}.
Furthermore, during the sampling process, diffusion models can be customized under the guidance of a classifier \cite{guided_diffusion}. This has rapidly become a research hotspot in the field of decision making. For example, the pioneering diffuser \cite{diffuser} method learns the dataset through the denoising process while interpreting the rewards in offline reinforcement learning as the classification probabilities of optimal trajectories to obtain a reward-guided diffusion model. The decision diffuser \cite{decision_diffuser} method expands the types of guidance by using classifier-free guidance to guide agents in generating trajectories that maximize rewards, satisfy trajectory constraints, and integrate existing skills. Moreover, some works leverage the guided diffusion approach, which modifies conditions to generate new data, for data augmentation. In policy-guided diffusion \cite{policy_guided_diffusion}, the discrepancy between the offline dataset of existing tasks and the target dataset of new tasks is used as guidance to generate more data to train a policy suitable for the new task. The adaptdiffuser \cite{adaptdiffuser} method uses a generator-discriminator architecture to generate data for task target generalization and has demonstrated its data augmentation advantages on robotic tasks. 

The method most similar to ours is the ReDiffuser \cite{rediffuser} method, which also evaluates reliability using RND; however, this indicator is not used to guide the diffusion model generation but only serves as a principle for selecting from multiple sampled trajectories. In contrast, we argue that introducing a reliability indicator during the guidance process is beneficial for directly generating more reliable trajectories to avoid hallucinations, and it can reduce the number of sampled trajectories, thereby improving inference efficiency.



\section{Method}

\begin{figure*}[thpb]
\centering
\includegraphics[width=0.9\textwidth]{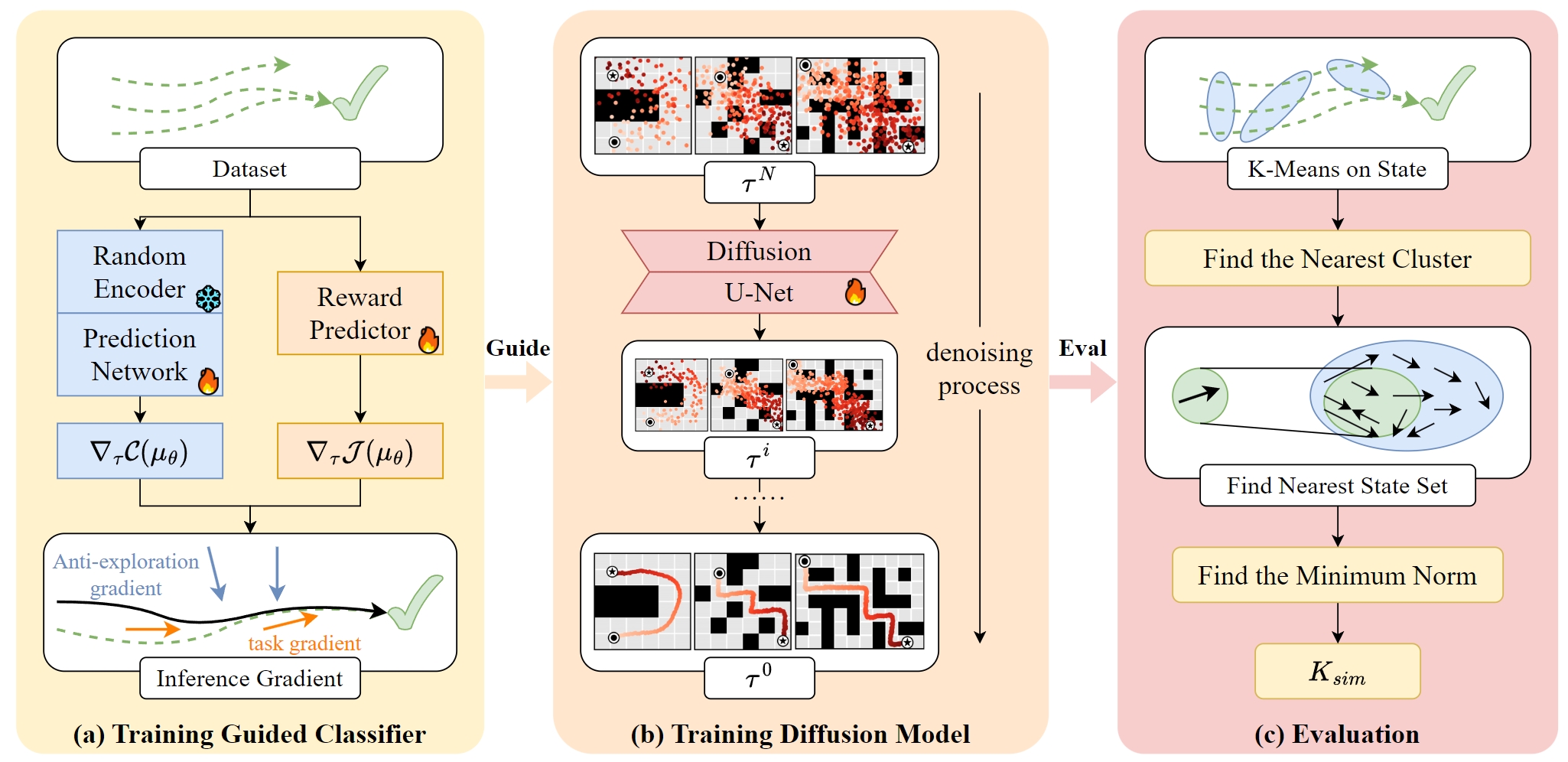}
\caption{
The overall framework of Curiosity-Diffuser is illustrated as follows. In part (a), we train the classifier for guidance. The dataset is used to train both the Prediction Network of the RND module and the Reward Predictor of the reward prediction module, where frozen network parameters are represented by a \textcolor{SkyBlue}{snowflake} symbol, and trainable network parameters are represented by a \textcolor{red}{flame} symbol. 
The gradients from these two modules yield two types of inference signals: one generated in the vicinity of the training data, and the other driving the model toward task completion. In part (b), a diffusion model is trained to mimic the policy embedded within the dataset, enabling the recovery of executable plans from noise samples. During inference, the classifier’s gradient is combined with the diffusion process to steer the generation of a specific policy. Finally, in part (c), we evaluate the generated results using the computationally efficient $K_{sim}$
metric, which verifies that Curiosity-Diffuser produces data with fewer hallucinations. In the accompanying figure, arrow positions represent states while arrow directions indicate actions, emphasizing that the search for the closest state–action pair prioritizes the state.
}
\label{main}
\end{figure*}


In this section, we first \textcolor{revision}{introduce} the diffusion model in decision-making and the conditional diffusion model with reward-preference characteristics. Based on this, we \textcolor{revision}{propose} the Curiosity-Diffuser method, which uses RND to evaluate trajectory performance and guide the diffusion model, thereby enhancing the reliability of the policy. Finally, we \textcolor{revision}{introduce} the K-Sim metric for closed-loop detection, which ensures that the Curiosity-Diffuser generates data closer to the training set, thus improving reliability. Fig. \ref{main} illustrates our main working approach.

\subsection{Diffusion Model for Decision}
\textcolor{revision}{The diffusion model treats the data training process as a forward diffusion process \( q(\tau^i \mid \tau^{i-1}) \) and an iterative denoising process \( p_\theta(\tau^{i-1} \mid \tau^i) \) \cite{diffusion_survey}}, where \( \tau \) represents the trajectory interwoven with states and actions \((s_0, a_0, s_1, a_1, \cdots, s_T, a_T)\), \( \theta \) denotes the learnable distribution parameters, \( t \) indicates the horizon of the trajectory, and \(i\in\{0,1,2,\cdots,N\}\) representing the diffusion steps of the diffusion process. The forward diffusion process introduces noise in multiple steps and eventually transforms the data distribution into a Gaussian distribution \( p(\tau^N) \), where each step transformation is typically a Gaussian distribution with variance increasing with the diffusion step \( i \).\textcolor{revision}{When the denoising process involves small steps, each step can be approximated as a Gaussian distribution with fixed covariance \cite{ddpm}} :


\begin{equation}
p_\theta(\tau^{i-1}\mid\tau^i)=\mathcal{N}(\tau^{i-1}\mid\mu_\theta(\tau^i,i),\Sigma^i)
\end{equation}
where \( \mu \) and \( \Sigma \) are the mean and covariance of the Gaussian distribution, with \( \mu \) encoded by the parameter \( \theta \).
Therefore, the parameters \( \theta \) can be updated by minimizing the error in noise prediction, enabling the model to generate data from noise as,

\begin{equation}
\theta^*=\arg\min_\theta-\mathbb{E}_{\tau^0}\left[\log p_\theta(\boldsymbol{\tau}^0)\right] 
\label{update_theta}
\end{equation}

Furthermore, by sampling multiple times from the Gaussian distribution $\textcolor{revision}{\tau^N}$ , diverse trajectory outputs can be generated. This makes it possible to perform multiple samplings during decision-making and select the optimal trajectory, enabling long sliding window execution similar to Model Predictive Control (MPC). Additionally, the diffusion model learns multiple modes of the data in a probabilistic manner, which allows it to learn decision-making behaviors from datasets with multiple action modes, while also possessing the capability to express various action modes. This extends the application of imitation learning to datasets collected by humans.

When it comes to robotic decision-making, we want the model to make the best possible decisions, rather than generating multiple styles of images as in the image generation field. In the context of offline RL, this can be seen as predicting the future behavior with the highest reward, while in the context of constrained trajectory generation, it can be viewed as generating trajectories that satisfy the constraint requirements. \textcolor{revision}{This type of problem can be modeled in the diffusion model as a conditional generation problem with a \textbf{guided diffusion model} \cite{guided_diffusion}} :
\begin{equation}
p_\theta(\tau^{i-1}\mid\tau^i,y(\tau^i)) \approx\mathcal{N}(\tau^{i-1};\mu_\theta+\alpha \Sigma g,\Sigma)
\label{guidance}
\end{equation}
where $y$ represents the category information about \( \tau^i \), such as whether the trajectory can complete the task, whether the trajectory meets the constraint conditions, and so on; \( \alpha \) represents the amplitude weight of the guidance information; \( g \) is the guidance information provided by the classifier, which directly affects the mean at each diffusion step. In offline RL, to maximize the expected reward, this guidance can be considered as the gradient of the cumulative reward derived from maximizing the probability of good trajectories, i.e.:

\begin{equation}
\begin{aligned}
g_1=& \nabla_{\tau}\log p_\phi(o_{0:T}\mid\tau)\mid_{\tau=\mu} \\
 =&\sum_{t=0}^{T}\gamma^{t}\nabla_{s_{t},a_{t}}\mathcal{R}(s_{t},a_{t})\mid_{(s_{t},a_{t})=\mu_{t}}=\nabla_{\tau}\mathcal{J}(\mu_{\theta})
\end{aligned}
\label{guide}
\end{equation}
\textcolor{revision}{where $p_\phi(o_{0:T}\mid\tau)$ is the likelihood that trajectory $\tau$ is optimal}, which can be decomposed into a series of terms proportional to \( \exp(\mathcal{R}(s_{t}, a_{t})) \). Therefore, it can be rewritten in the summation form, with \( \gamma^t \) as the discount factor, and the cumulative reward is expressed as \( \mathcal{J} \). 

This brings about a beneficial property: Equations \ref{update_theta} and \ref{guide} can be learned independently. Equation \ref{update_theta} uses a mix of good and bad diverse trajectories to update the network parameters \( \theta \) for policy learning, while Equation \ref{guide} trains the classifier parameters \( \phi \) based solely on the category of trajectory data, without requiring policy learning. During inference, the gradient of the classifier is computed to obtain the guiding knowledge for the diffusion model. This flexible approach also allows us to introduce more guidance information. In the next section, we will introduce how we use RND as a representative model evaluation to guide the diffusion model in generating behaviors that are more familiar to itself, thereby improving the credibility of the generated outputs.

\subsection{RND-based Model Evaluation}

Model evaluation refers to detecting the effectiveness of the model's output in some way. In open-loop and non-interpretable systems like neural network models, evaluating the quality of their outputs becomes particularly crucial. Current methods for evaluating neural networks include analyzing the distribution of neuron activation levels, examining attention weights in attention mechanisms, and introducing another model to form a closed loop. In this work, we focus on introducing an evaluation model to assess the results, thus drawing attention to the \textcolor{revision}{RND method \cite{rnd}} from the reinforcement learning field as a method of model evaluation. Specifically, RND introduces a fixed-parameter random encoder \( f_{\xi}(\tau) \) and a learnable prediction network \( f_{\eta}(\tau) \), both of which take training data as input. Therefore, \( \eta \) can be learned as follows:

\begin{equation}
\eta =\arg\min_\eta \mathbb{E}_{\tau^i} \| f_{\eta}(\tau^i) - f_{\xi}(\tau^i) \|_2^2
\label{K_sim}
\end{equation}

The reason for using \( \textcolor{revision}{\tau^i} \) in the diffusion steps is that during inference, the diffusion model generates semi-mature trajectories, which also need to be learned by the prediction network to ensure that effective guidance is produced at each diffusion step. Noticing that curiosity value \( \| f_{\eta}(\tau^i) - f_{\xi}(\tau^i) \|_2^2 \) is inversely proportional to the probability of generating a reliable trajectory, i.e., the smaller the absolute difference between the two networks, the more reliable \( \tau^i \) is. Therefore, the RND-based guidance can be expressed as:

\begin{equation}
\begin{aligned}
g_2&= \nabla_{\tau}\log p_{\eta,\xi}(r\mid\tau^i)\mid_{\tau^i=\mu^i} \\
 &=-\nabla_{\tau} \| f_{\eta}(\tau^i) - f_{\xi}(\tau^i) \|^2_2=\nabla_{\tau}\mathcal{C}(\mu_{\theta})
\end{aligned}
\end{equation}
where \( p_{\eta,\xi}(r \mid \tau^i) \) represents the probability that \( \tau^i \) is reliable.

And to prevent the model from only pursuing familiar trajectories without completing the task, we made two improvements. On one hand, we selected only successful data to train the RND prediction network, ensuring that familiar data can also complete the task. On the other hand, by combining RND guidance with simple reward-based guidance, we ensure that the generated samples balance both reliability and success. Therefore, the guidance term can be expressed as:

\begin{equation}
g=g_1+\lambda g_2
\label{guide_ratio}
\end{equation}
where \( \lambda \) represents the weighting coefficient. This term will be used in Equation \ref{guidance} to guide the diffusion model in generating plans with comprehensive performance.

The RND method is versatile, and the guidance of the diffuser model also exhibits good scalability. Therefore, the RND-based model evaluation and imitation learning generation guidance are universally applicable and suitable for a wide variety of tasks.

\subsection{Similarity Measurement}

To evaluate the effect of introducing RND guidance, we also need a closed-loop test to assess the similarity between the trajectories in imitation learning tests and the imitation learning training set. Unlike approaches such as mAP in deep learning, the error in decision-making problems depends on the action differences in similar states, meaning the similarity of states has higher priority than the actions. Therefore, evaluation of decision-making problems cannot be directly performed by averaging the error across all state-action pairs \((s_k, a_k)\) in the training set, but instead needs to be evaluated in stages. Searching for similar states across the entire training set would result in prohibitively low computational efficiency. Thus, we have designed a computationally efficient similarity calculation method based on K-Means clustering, called K-Sim, denoted as $K_{sim}$.

Specifically, we first preprocess the training set using the K-Means method based on states \( s_j \), dividing it into \( m \) K-Means clusters: $\{(s_k, a_k)\}_{1,...,j,...,m} $
where \( k \) is the index of the data pair, and \( j \) is the index of the K-Means cluster. Furthermore, for the \( i \)-th decision during testing \( (s_i, \pi(s_i)) \), we find the cluster \( \{(s_k, a_k)\}_j \) that is closest to \( s_i \) by searching through the center of each K-Means cluster. Then, we traverse this cluster and select a small set of states that are most similar to \( s_i \) (avoid greedy state selection, which may result in actions that are not close). From this set, we identify the action \( a_k \) that is most similar to \( \pi(s_i) \), and use the corresponding state-action pair \( (s_k, a_k) \) as the nearest pair. Finally, we measure the similarity between \( (s_k, a_k) \) and \( (s_i, \pi(s_i)) \) using the norm distance between them. \textcolor{revision}{Before computing K-Sim, states and actions are normalized using the statistics of the training dataset, so \(s\) and \(a\) in the following calculation denote normalized vectors. This preprocessing reduces the sensitivity of the metric to the high dimensions and numerical scales of robot state-action spaces.}
\textcolor{revision}{\begin{equation}
K_{sim}= \dfrac{1}{n} \sum_i^n  \min(1,\frac{\gamma\sqrt{d}}{\| (s_i,\pi(s_i)),(s_{ki},a_{ki})\|_2}) 
\end{equation}}
where \( n \) represents the number of time steps during testing, $d$ represents the dimensions of state and action, and \( (s_{ki}, a_{ki}) \) represents the nearest pair found for each \( (s_i, \pi(s_i)) \) in the training set. \textcolor{revision}{The $\sqrt{d}$ is used to normalize the effect of the vector dimension on the L2 norm, and the coefficient \( \gamma \) is used to control the tolerance of the bounded similarity measurement rather than to provide a universal cross-environment scale. A larger \( \gamma \) makes the metric more permissive to state-action differences.} Therefore, the closer the $K_{sim}$ is to 1, the more similar the behavior during inference is to the training data. A score closer to 0 indicates that the behavior during inference exceeds the coverage of the training data, making it more likely to perform poorly.

The advantage of K-Sim lies in its balance between computational efficiency and accuracy. A brute-force approach would require computations proportional to the size of the training dataset multiplied by the number of test trajectory points, which is highly inefficient. In contrast, the computational complexity of the K-Means algorithm is proportional to the size of the training dataset, the number of clusters, and the number of iterations, making it computationally feasible and capable of handling $10^6$
data points within a minute. Furthermore, with the preprocessing done by the K-Means algorithm, each test trajectory point only needs to traverse a specific cluster, significantly reducing the computational load and enabling fast evaluation.


\section{Experiments}
With our experimental evaluations, we aim to answer the
following questions:

\begin{itemize}
\item Does our proposed curiosity guidance improve the performance of diffusion-based robotic policies?
\item Does the curiosity-diffuser effectively reduce model hallucinations?
\item Is the policy's performance sensitive to the guidance mixing weight $\lambda$?
\item Can our method deliver superior performance in real-world robot task?
\end{itemize}

We conduct experimental evaluations in both simulation and
the real world.
We provide comprehensive answers to the first two questions through comparisons with similar methods in simulation and analysis of specific scenarios, and address the third question via multi-parameter comparisons. Finally, the real-world manipulation tests are used to validate the effectiveness of policy deployment regarding the last question.

\begin{figure}[thpb]
  \centering
  \includegraphics[width=0.45\textwidth]{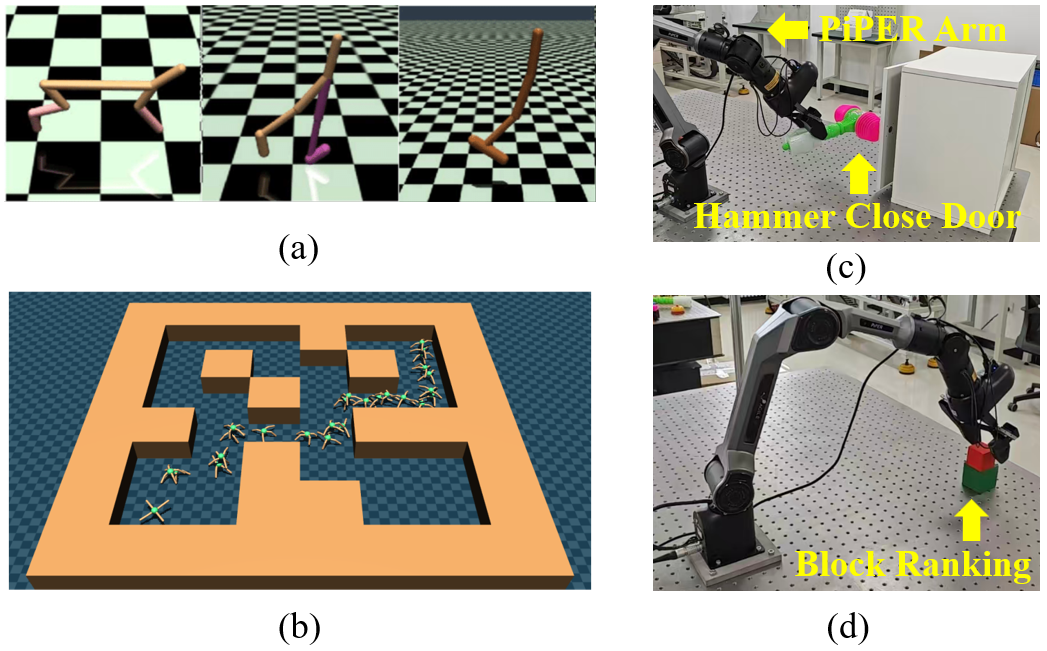}
  \caption{(a) HalfCheetah, Walker2d and Hopper in Gym-Mujoco of D4RL dataset. (b) AntMaze in Minari dataset. (c) \& (d) Hammer Close Door and Block Ranking with PiPER manipulator of AgileX Robotics in real-world.}
  \label{experiment}
\end{figure}

For the simulation, we chose the widely used D4RL \cite{d4rl} library's MuJoCo and Minari \cite{minari} library's AntMaze environments as our simulation validation environments, which are shown in Fig. \ref{experiment} (a) and (b). 
The MuJoCo environment consists of three popular offline RL motion tasks (HalfCheetah, Hopper, and Walker2d), which require controlling legged MuJoCo robots to move forward while minimizing energy consumption. The D4RL benchmark provides three different quality levels of offline datasets: "Medium" contains demonstrations with medium-level performance; "Medium Replay" contains data with lower performance than "Medium" that was observed during training; "Medium Expert" refers to a combination of "Medium" and "Expert" level performances in equal proportions. 
The AntMaze environment requires controlling an eight-degree-of-freedom "ant" quadruped robot to complete a maze navigation task. In the dataset, the robot only receives a sparse reward when it reaches the goal, making it a challenging decision-making task. The "Medium" and "Large" configurations determine the size of the maze map, where in the Play group, the robot's starting point is random and the target point is fixed; in the Diverse group, both the robot's starting point and the target point are random. 
Noticed the D4RL library uses an incorrect policy when collecting dataset on the Antmaze task, we choose the Minari library, which corrects this issue, as the benchmark for evaluating the Antmaze task.

To validate the effectiveness of our algorithm in real-world scenarios, we selected and adapted two manipulation tasks from the RoboTwin benchmark \cite{robotwin}: \textit{Hammer Close Door} and \textit{Block Ranking}.
The specifics of the tasks are defined as follows:
\begin{itemize}
\item 
\textbf{Hammer Close Door}: The robotic arm is required to grasp a hammer and subsequently use it to push a cabinet door closed.
\item \textbf{Block Ranking}: This task involves precise pick-and-place operations, where the robot must grasp a target red block on the table and stack it onto a specific green block.
\end{itemize}
For data collection, we utilized the AgileX \textbf{Pika} handheld teleoperation device to control a \textbf{PiPER} robotic arm. And we recorded synchronized RGB visual data from a multi-view setup, consisting of two static cameras and one eye-in-hand camera.
We collected a dataset of 100 expert trajectories for each task to ensure a fair comparison with state-of-the-art imitation learning methods. 
To introduce visual complexity and test robustness, task-irrelevant distractor objects (e.g., blue blocks) are placed within the workspace.
Crucially, to ensure data diversity, the initial positions of all objects—including both task-relevant items and distractors—were randomized during the data collection process. 
{\color{revision}The robot arm was initialized from a fixed pose during data collection, while the object placement was randomized across demonstrations. During evaluation, the object placement remained randomized, and an additional small perturbation was applied to the robot's initial pose. Thus, the real-world evaluation mainly tests deployment perturbations and mild distribution shifts induced by robot initial-pose changes and potentially unseen pose-object combinations, rather than arbitrary OOD disturbances.}
During the evaluation phase, the trained policy processes these visual inputs to generate real-time motion commands for the robotic arm, enabling autonomous task execution.

\subsection{Simulation Performance Evaluation}

\begin{table*}[htbp]
\begin{center}
\tiny
\setlength{\tabcolsep}{4pt}
\begin{minipage}{1.0\textwidth}
\caption{Evaluation results of Mujoco environments in D4RL dataset. In the table, our method Curiosity-Diffuser is labeled as \textbf{Ours}. We report the mean scores over 150 planning episodes, and the top-3 scores per task are bolded.}
\begin{tabularx}{1.0\textwidth}{@{}ccccccccccccc@{}}
\toprule
\textbf{Dataset}& \textbf{Env.} & \textbf{BC}  & \textbf{MOPO} & \textbf{MBOP}  & \textbf{SynthER}  & \textbf{MOReL}&\textbf{DT} & \textcolor{revision}{\textbf{ReDiffuser}} & \textbf{Diffuser} & \textbf{IQL} & \textbf{CQL} & \textbf{Ours}  \\ \midrule
\multirow{3}{*}{Medium-Expert}  & HalfCheetah & 55.2   & 63.3  & \textbf{105.9}  & 86.6 & 53.3 & 86.8 & 82.7 & 86.3  & 86.7 & \textbf{91.6} & \textbf{88.3}   \\
 & Hopper& 52.5  & 23.7  & 55.1 & 94.7 & 
 \textbf{108.7}& \textbf{107.6} & 102.6 & 105.2  & 91.5 & 105.4 & \textbf{108.3} \\
 & Walker2d   & 107.5& 44.6& 70.2 & \textbf{109.7}  & 95.6& 108.1 & 107.7 & 106.4 & \textbf{109.6}  & \textbf{108.8} & 107.6 \\ 
 \midrule
 
\multirow{3}{*}{Medium}  & HalfCheetah & 42.6 & 42.3 & 44.6 & \textbf{46.5}  & 42.1 & 42.6 & 43.9 & \textbf{45.2} & \textbf{47.4} & 44.0 & \textbf{45.2}  \\
 & Hopper  & 52.9 & 28.0& 48.8 & 58.2 & \textbf{95.4} & 67.6 & \textbf{93.6} & \textbf{94.9} & 66.3 & 58.5 & 92.5 \\ 
& Walker2d & 75.3 & 17.8 & 41.0 & \textbf{77.6} & \textbf{77.8} & 74.0 & 75.1 & 77.1 & \textbf{78.3} & 72.5 & 77.4 \\ \midrule
\multirow{3}{*}{Medium-Replay}   & HalfCheetah   & 36.6 & \textbf{53.1}& 42.3 & 28.0 & 40.2 & 36.6  & 35.2 & 33.5  & \textbf{44.2} & \textbf{45.5}   & 35.2 \\ 
& Hopper     & 18.1  & 67.5&12.4  & 24.7 & \textbf{93.6} & 82.7  & 93.4 & 91.6 & \textbf{94.7} & \textbf{95.0} & 92.4 \\ 
& Walker2d     & 26.0  & 39.0 & 9.7  & 51.4 & 49.8& \textbf{66.6} & 52.3 & 53.1 & \textbf{73.9}  & \textbf{77.0} & 54.4 \\ 
\midrule
\multirow{1}{*}{Average}    &        & 51.9 & 42.1 & 47.8 & 64.2 & 72.9 & 74.7 & 76.3 & \textbf{77.0} & \textbf{77.0}  & \textbf{77.6} & \textbf{77.9} (best) \\ \botrule
\end{tabularx}
\label{tab:mujoco_result}
\end{minipage}
\end{center}
\end{table*}

\begin{table*}[htbp]
\begin{center}
\footnotesize
\begin{minipage}{1.0\textwidth}
\caption{Evaluation results of AntMaze environments in Minari dataset. We report the mean scores over 150 planning episodes, and the top scores per task are bolded.}\label{tab:antmaze_result}
\begin{tabularx}{1.0\textwidth}{@{}ccccccccccc@{}}
\toprule
\textbf{Dataset}& \textbf{Environment} & \textbf{BC}  & \textbf{DD} & \textbf{AdaptDiffuser}  & \textbf{Diffuser} & \textcolor{revision}{\textbf{ReDiffuser}} &\textbf{DQL} & \textbf{Curiosity-Diffuser (Ours)}  \\ \midrule
\multirow{2}{*}{Play}   & Antmaze-Medium     & 0.0    & 8.0 & 4.6  & 1.3 & 3.3 & 4.0  & \textbf{58.6}  \\ 
& Antmaze-Large   & 0.0    &0.0& 0.0   & 0.7    & 1.3 &0.0&\textbf{34.0}  \\ 
\multirow{2}{*}{Diverse} & Antmaze-Medium     & 0.8   &4.0& 2.7  & 0.7  & 3.3 & 0.0  & \textbf{24.0}  \\ 
& Antmaze-Large   & 0.0 &0.0&0.0 & 2.0  & 1.3 & 0.0  & \textbf{42.0}  \\ \midrule
\multirow{1}{*}{Average} &     & 0.2 &3.0& 1.8 & 1.2 & 2.3 & 1.0 &  \textbf{39.7} \\ \botrule
\end{tabularx}
\end{minipage}
\end{center}
\end{table*}

{\color{revision}Table \ref{tab:mujoco_result} and \ref{tab:antmaze_result} show the comparison of our simulation results with BC \cite{BC}, MOPO \cite{MOPO}, MBOP \cite{MBOP}, SynthER \cite{SynthER}, MOReL\cite{MoRel}, Decision Transformer (DT) \cite{decision_transformer}, Implicit Q-Learning (IQL) \cite{IQL}, Conservative Q-Learning (CQL) \cite{CQL}, Decision Diffuser (DD) \cite{decision_diffuser}, AdaptDiffuser \cite{adaptdiffuser}, Diffusion Q-learning (DQL) \cite{diffusion_qlearning}, Diffuser \cite{diffuser} and ReDiffuser \cite{rediffuser}. These baselines cover imitation learning, offline reinforcement learning, diffusion-based data synthesis, reward guided diffusion planning and curiosity-driven planning, thereby providing a broad comparison for evaluating the effect of curiosity guidance.}

{\color{revision}Our implementation is built on the CleanDiffuser framework \cite{cleandiffuser}, an open-source modular library designed for diffusion models in decision-making tasks. CleanDiffuser provides reusable components for diffusion models, network architectures, conditioning mechanisms, guided sampling, environment interfaces, and training pipelines. We use it as the base implementation framework for the diffusion-policy training and baseline reproduction, and add the proposed RND-based curiosity guidance and reward guidance on top of its guided diffusion pipeline. The main hyperparameters of our simulation experiments are shown in Table \ref{tab:simulation_hyperparameters}. The subsequent section provides a more detailed analysis of the simulation results.}


\subsubsection{MuJoCo Task}

In the MuJoCo tasks, Curiosity-Diffuser achieves a modest improvement in the average score, although several baselines still perform better on particular tasks. The performance improvement is most notable in the Medium-Expert group. Through analyzing the differences in the content of the three datasets, we found that only the Medium-Expert group contains high-quality expert demonstrations, which leads to better behaviors being guided by Random Network Distillation. 
{\color{revision}ReDiffuser, which also uses RND-based confidence estimation, obtains competitive results in several MuJoCo settings, further suggesting that reliability estimation is useful for diffusion-based decision making. However, ReDiffuser mainly uses the RND signal to select among generated candidate trajectories, whereas Curiosity-Diffuser introduces RND guidance during the denoising process.} 

In contrast, the data quality in the Medium and Medium-Replay groups is lower, and thus the behaviors guided by RND guidance may have a negative impact on task completion.
For example, in the Medium-Replay group, the performance of HalfCheetah and Walker2d tasks under curiosity guidance or post selection remains lower than that of IQL and CQL methods. This indicates that our approach exhibits some sensitivity to data quality. 
{\color{revision}Overall, Curiosity-Diffuser achieves the best average performance, but the MuJoCo results should be interpreted as a modest and data-dependent improvement rather than a uniform advantage over all baselines.}

\begin{table*}[htbp]
\color{revision}
\centering
\footnotesize
\caption{Key hyperparameters used for Diffuser and Curiosity-Diffuser in the simulation experiments.}
\begin{tabularx}{1.0\textwidth}{@{}p{0.16\textwidth}p{0.24\textwidth}X X@{}}
\toprule
\textbf{Method} & \textbf{Hyperparameters} & \textbf{Task-MuJoCo} & \textbf{Task-AntMaze} \\
\midrule
\multirow{11}{*}{Diffuser} & Diffusion solver & DDPM & DDPM \\
& U-Net model dim & 32 & 64 \\
& Diffusion training steps & 20 & 20 \\
& Sampling steps & 20 & 20 \\
& Planning horizon & 32 & 64 \\
& Reward classifier net & HalfJannerUNet1d & HalfJannerUNet1d \\
& Guidance weight $\alpha$ & $[1\times10^{-5},1\times10^{-3}]$ & $1\times10^{-3}$\\
& Guidance scale schedule & Constant over diffusion steps & Constant over diffusion steps \\
& Diffusion training steps & $1\times10^6$ & $1\times10^6$ \\
& Classifier training steps & $1\times10^6$ & $1\times10^6$ \\
& Batch size & 64 & 64 \\
\midrule
\multirow{4}{*}{Curiosity-Diffuser} & Curiosity weight ($\lambda$) & $1\times10^6$ & $1\times10^6$ \\
& RND classifier net & HalfJannerUNet1d & HalfJannerUNet1d \\
& RND classifier training steps  & $3\times10^5$ & $3\times10^5$  \\
& batch size & 1024 &1024\\
\botrule
\end{tabularx}
\label{tab:simulation_hyperparameters}
\end{table*}

\subsubsection{AntMaze Task} \label{antmaze_analysis}

In the AntMaze task, the dataset provides only sparse rewards that indicate whether the goal has been reached.
We observed that BC, DD, AdaptDiffuser, Diffuser, ReDiffuser, and DQL almost completely fail to complete the task, while our model achieves impressive performance. Under sparse reward conditions, since rewards at each step carry little information, RND guidance becomes even more critical. \textcolor{revision}{The comparison with ReDiffuser further indicates that using RND only to select among generated candidate trajectories is insufficient for this long-horizon navigation task; by contrast, Curiosity-Diffuser uses the RND signal during the denoising process, allowing the generation procedure itself to be guided toward more familiar and reliable trajectory regions.} Furthermore, we observe that the Large group is more challenging than the Medium group. This is because longer task completion processes are more susceptible to the accumulation of imitation learning errors. For instance, methods such as DD and AdaptDiffuser achieve 0\% success in the Large configuration, whereas they attain modest success rates in the Medium group. Consequently, it is understandable that the Curiosity-Diffuser method achieves a higher success rate in the Medium configuration of the Play group compared to the Large configuration.
But in the Diverse group, our success rates remain relatively low. This can be attributed to the randomization of starting points and goal locations, which introduces greater diversity in the motion data within the dataset. As a result, curiosity does not always effectively guide the current policy toward data well-suited to the specific target. Nevertheless, reducing inference hallucinations still proves beneficial, yielding a noticeable improvement in success rates.

\begin{figure}[thpb]
  \centering
  \includegraphics[width=0.48\textwidth]{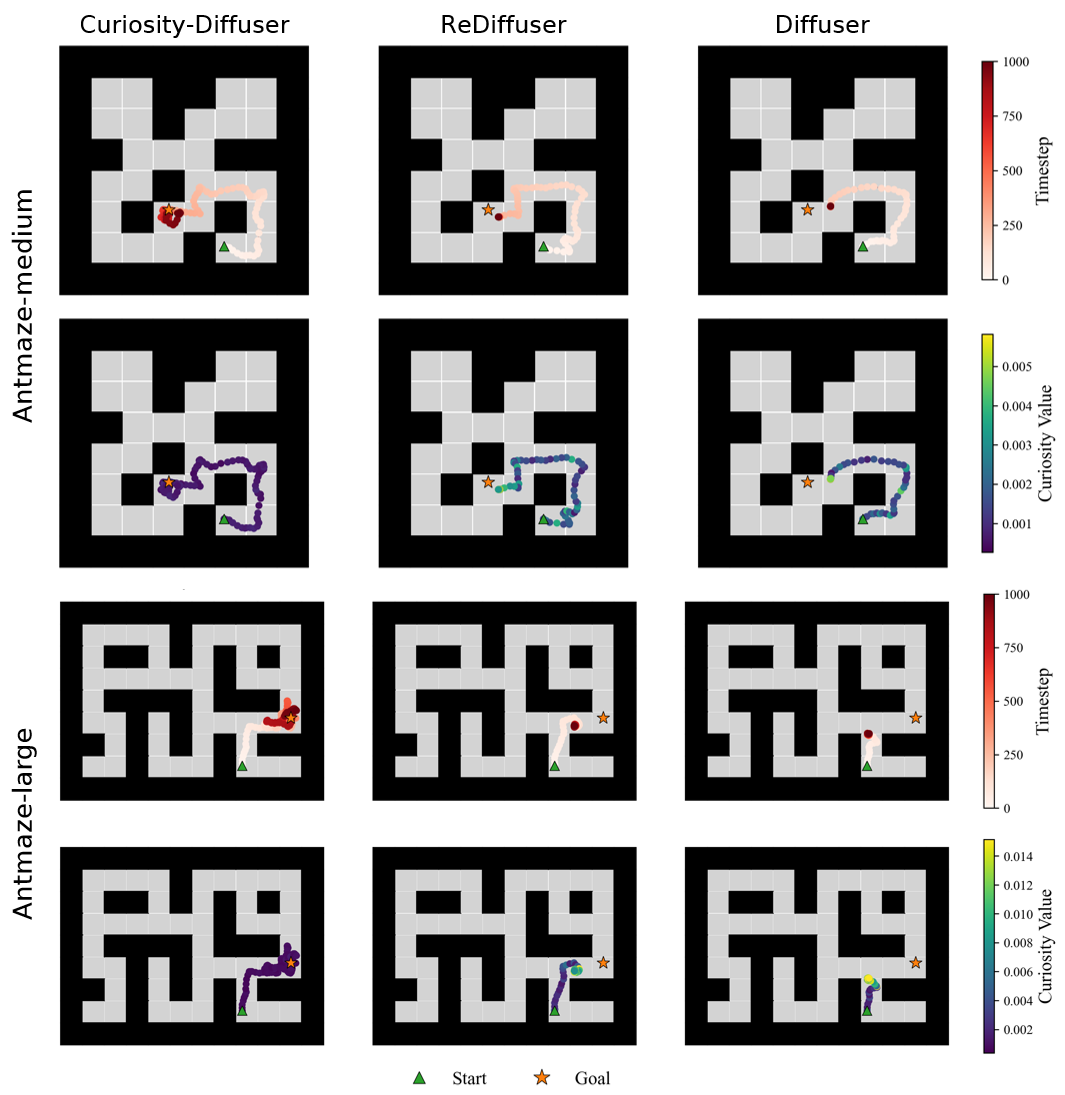}
  \caption{\textcolor{revision}{Visualization of the trajectories generated by three methods in the medium and large AntMaze tasks. The start and goal positions are indicated by a green triangle and an orange star, respectively. The robot’s center-of-mass trajectory is shown as a curve ranging from light red to dark red. The curiosity values $\mathcal{C}$ along the trajectory are represented by a blue-to-green-to-yellow color gradient, where colors closer to deep blue indicate lower curiosity.}}
  \label{maze_path}
\end{figure}

In the first column of Fig. \ref{maze_path}, the RND-guided robot is able to move toward the goal with lower curiosity. Even when hesitation occurs with the robot stepping in place along the path, it can re-enter the familiar trajectory with the combined guidance of RND and rewards.
\textcolor{revision}{In contrast, in the second and third columns, when only reward guidance is used, the robot’s trajectories become more unstable, indicating that the model produces more hallucinations during inference. In both cases, the failures are caused by the robot entering states with high curiosity values, where effective reasoning can no longer be performed, causing it to become stuck in place until the end of the episode. This suggests a close relationship between policy hallucinations and task failure. 
Notably, ReDiffuser exhibits lower curiosity values than Diffuser, indicating that curiosity-based trajectory selection is effective to some extent and generally enables the robot to move closer to the task goal. However, when the diffusion model itself fails to generate any effective actions, ReDiffuser also becomes stuck in place. This demonstrates that curiosity-based guidance can better exploit the knowledge of the diffusion model and produce actions with fewer hallucinations.}


\subsection{K-Sim Evaluations}

\textcolor{revision}{We report the K-Sim scores of Diffuser, ReDiffuser, and Curiosity-Diffuser calculated using Equation \ref{K_sim} in Table \ref{tab:K_sim}. A score closer to 1.0 indicates that the closed-loop state-action outputs are more similar to the training dataset, while a lower score suggests a larger deviation from the training support and a higher risk of unreliable generation.}
For each group, the K-Means clusters are set to 50 and the size of the Nearest State Set is set to 50. \textcolor{revision}{The state-action vectors are normalized with training-set statistics before computing K-Sim, and \( \gamma \) is set to 0.5.} Under this configuration, the calculation of each indicator can be completed in about \textcolor{revision}{1 minute}, which is better than traversing the entire training dataset.

\begin{table}[htbp]
\color{revision}
\begin{center}
\footnotesize
\setlength{\tabcolsep}{3pt}
\begin{minipage}{0.48\textwidth}
\caption{K-Sim similarity ($\uparrow$) evaluation in the AntMaze environment. Curiosity-Diffuser is labeled as \textbf{Ours}}
\begin{tabular}{ccccc}
\toprule
\textbf{Dataset}& \textbf{Environment} & \textbf{Diffuser} & \textbf{ReDiffuser}& \textbf{Ours} \\
\midrule
\multirow{2}{*}{Play}   & Antmaze-Medium  & 0.767  & 0.824 &\textbf{0.868} \\ 
& Antmaze-Large & 0.726 & \textbf{0.787}  & \textbf{0.787} \\ 
\midrule
\multirow{2}{*}{Diverse} & Antmaze-Medium     & 0.757  & 0.752 & \textbf{0.824}    \\ 
& Antmaze-Large     & \textbf{0.893}  & 0.817  & 0.829    \\
\midrule
Average & & 0.785&0.795&\textbf{0.827}\\
\botrule
\end{tabular}
\label{tab:K_sim}
\end{minipage}
\end{center}
\end{table}

\textcolor{revision}{The data show that Curiosity-Diffuser obtains the highest average K-Sim score, 0.827, compared with 0.785 for Diffuser and 0.795 for ReDiffuser. Curiosity-Diffuser also outperforms ReDiffuser in three of the four AntMaze settings and achieves the same score in Antmaze-Large-Play. This suggests that introducing the RND signal into the denoising process helps keep the closed-loop behavior closer to the training dataset regions compared with using RND only to select among generated candidate trajectories. Meanwhile, ReDiffuser improves the average score over Diffuser, indicating that RND-based reliability estimation is also useful for trajectory selection.}

\textcolor{revision}{Nevertheless, K-Sim should be interpreted together with task success and qualitative trajectory evidence, rather than as a standalone proof of eliminating hallucinations. For example, Diffuser obtains the highest K-Sim score in Antmaze-Large-Diverse, and Fig. \ref{maze_path} shows that ReDiffuser can become stuck near the goal. In such cases, some failed or stuck states may still be close to existing training samples in Euclidean distance, which can keep the K-Sim score relatively high even when the task is not completed. Overall, the K-Sim results provide complementary closed-loop evidence that Curiosity-Diffuser tends to keep generated behaviors closer to the demonstrated data support, while the final reliability conclusion should be drawn jointly from K-Sim and task success.}

\subsection{Sensitivity Analysis of Curiosity Guidance Weight}

\begin{figure}[thpb]
  \centering
  \includegraphics[width=0.48\textwidth]{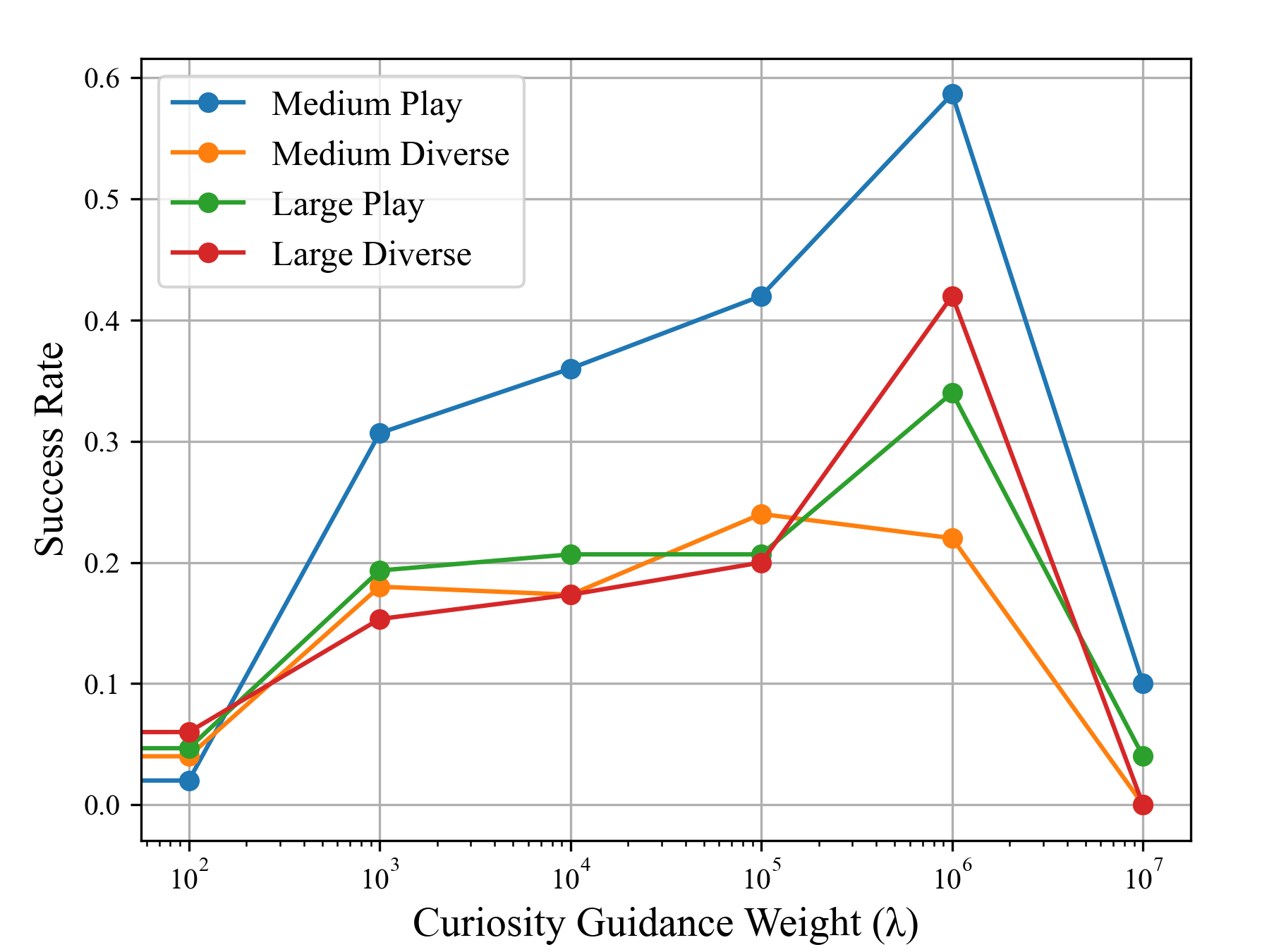}
  \caption{Ablation experiment of curiosity guided weight $\lambda$ on Antmaze tasks, where the dots represent the sampling points of $\lambda$ and are connected with curves to indicate the trend. For $0\leq\lambda<10^2$, curiosity has almost no impact on the diffusion inference process. The success rates are nearly the same as when $\lambda=10^2$, so these cases are not plotted.}
  \label{ratio_figure}
\end{figure}

We observed some interesting details in the experiment: the weighting factor \( \lambda \) in Equation \ref{guide_ratio} has a non-monotonic effect on task performance. We plotted the experimental results with various values of \( \lambda \) for the AntMaze task in Fig. \ref{ratio_figure} and found that as \( \lambda \) increases from 0, the task performance initially improves. However, after reaching a certain threshold, performance begins to decline, potentially even worse than the case when \( \lambda = 0 \). This suggests that excessive reliance on RND-based guidance may be detrimental to task completion. 

This situation highlights the necessity of both types of guidance. If only reward-based guidance is used, the model is prone to generating hallucinations during inference due to OOD generalization, which leads to execution failure. On the other hand, if only RND-based guidance is used, the model will tend to focus on the neighborhood of the dataset, without being able to complete the task. The combination of both types of guidance forms a comprehensive guiding mechanism, which encourages the generation of inference gradients that balance the goals of task completion and minimizing hallucinations, as shown in (a) of Fig. \ref{main}, thus resulting in good behavior.

\subsection{Real-World Evaluations}

To rigorously evaluate the performance of our proposed method in real-world manipulation tasks, we conducted comparative experiments against several strong baselines. We selected two state-of-the-art single-task imitation learning methods: Action Chunking Transformer (ACT) \cite{aloha} and Diffusion Policy (DP) \cite{diffusion_policy}. Furthermore, we extended the vanilla DP with classifier guidance to construct a baseline Reward-Guided Diffuser and our Curiosity-Diffuser, and we set $\lambda=10^2$ for our method.
All models were trained using 100 successful trajectories and evaluated on 20 trials. For the guided methods (Reward-Guided Diffuser and Curiosity-Diffuser), we employed a strict sparse reward setting where a sparse terminal reward is provided only at the end of the trajectory.

The quantitative results of the real-world evaluations are summarized in Table \ref{tab:real_metric}. We observe that under the limited data regime, the standard imitation learning baselines, ACT and DP, exhibited mediocre performance, achieving an average success rate of only around 50\%. These methods proved sensitive to minor variations in environmental setups, such as changes in the robot's initial pose, leading to significant performance degradation. In contrast, the result of Diffuser demonstrated that incorporating reward guidance effectively improves the success rate to 75\%. Qualitatively, we found that reward guidance serves to accelerate the policy's motion, mitigating the issue of stationary "hovering" in generated trajectories. Most notably, our proposed Curiosity-Diffuser outperformed all baselines with an average success rate of 85\%. \textcolor{revision}{This result indicates improved stability under the mild deployment perturbations in our evaluation, including small robot initial-pose changes and randomized object placements.}



\begin{table}[htbp]
\begin{center}
\footnotesize
\begin{minipage}{0.5\textwidth}
\caption{Evaluation results of Real-World Manipulation Tasks. Curiosity-Diffuser is labeled as \textbf{Ours}. The performance is evaluated using success rate, averaged over 20 trials.}
\begin{tabular}{@{}ccccc}
\toprule

\textbf{Task}& \textbf{ACT} & \textbf{DP} & \textbf{Diffuser} & \textbf{Ours} \\
\midrule
Hammer Close Door & 50\%  & 60\% & 75\%  & \textbf{90\%} \\  

Block Ranking & 45\%  & 55\% & 75\%  & \textbf{80\%} \\ 

\midrule
Average & 47.5\% & 57.5\% & 75\% & \textbf{85\%} \\
\botrule
\end{tabular}
\label{tab:real_metric}
\end{minipage}
\end{center}
\end{table}

\section{Conclusions}

We propose Curiosity-Diffuser, a decision-making method based on diffusion models that enhances reliability through curiosity-based evaluation. By leveraging the RND curiosity approach to assess data novelty, we incorporate curiosity guidance during the inference process of the conditional diffusion model to suppress OOD behaviors. Thus, Our method notably improves the performance of diffusion models in existing decision tasks. We validated our approach using several widely adopted offline RL simulation tasks and real-world manipulation tasks, and we introduced the computationally efficient K-Sim metric to quantify improvements in decision-making reliability.

Further improving the performance of model evaluation methods is a potential area for future work. In testing with some simple examples, we found that the RND method still make errors sometimes. 
Inspired by recent advancements in mechanistic interpretability, such as sparse autoencoders and neural circuits, we recognize that analyzing internal neuronal activation patterns offers deeper insights into model behavior. Consequently, we plan to leverage these techniques to develop more robust OOD detection mechanisms in our future research.

\section*{Acknowledgments}

This work was supported in part by Guangdong Major Project of Basic and Applied Basic Research under Grant 2023B0303000016, and the National Natural Science Foundation of China under Grant 92370123 and 62273280.

\section*{Declarations of Conflict of Interest}
The authors declared that they have no conflicts of interest to this work.






\bibliography{sn-bib}

\bigskip

\begin{figure}[H]%
\centering
\includegraphics[width=0.15\textwidth]{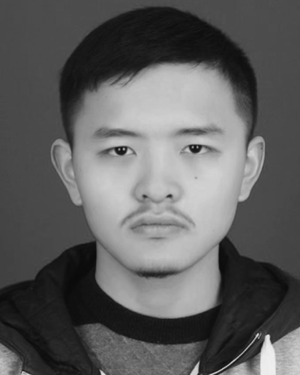}
\end{figure}

\noindent{\bf Zihao Liu }\quad received the B.S. degree in aerospace engineering from the School of Astronautics, Northwestern Polytechnical University, Xi’an, China, in 2023. He is currently pursuing the Ph.D. degree in control science and engineering with the School of Astronautics, Northwestern Polytechnical University.

His research interests include embodied artificial intelligence, diffusion models, and robotic manipulation.

E-mail: liuzihao@mail.nwpu.edu.cn

ORCID iD: 0009-0008-2637-4824

\begin{figure}[H]%
\centering
\includegraphics[width=0.15\textwidth]{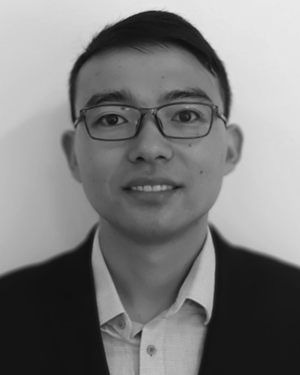}
\end{figure}

\noindent{\bf Xing Liu }\quad received the Ph.D. degree from the Shaanxi Key Laboratory of Intelligent Robots, Xi’an Jiaotong University, Xi’an, China, in December 2019. He was a Visiting Ph.D. Stu dent with the Social Robotics Laboratory, National University of Singapore, Singapore. He is now a Pro fessor with the School of Astronautics, Northwestern Polytechnical University. 

His research focuses on robot manipulation skill learning, reinforcement learning, and human–machine hybrid intelligence.

E-mail: xingliu@nwpu.edu.cn (Corresponding author)

ORCID iD: 0000-0002-5327-4908

\begin{figure}[H]%
\centering
\includegraphics[width=0.15\textwidth]{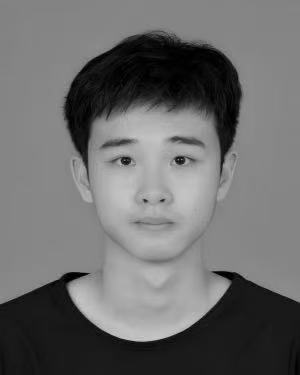}
\end{figure}

\noindent{\bf Yuhang Dong }\quad received the B.S. degree in Mechatronic Engineering from the College of Engineering, China Agricultural University, Beijing, China. He is currently pursuing the M.Eng. degree in Control Engineering with the Polytechnic Institute of Zhejiang University, Hangzhou, China. 

His current research interests include embodied intelligence, robotic manipulation, world models, and multimodal large language models.

E-mail: 22360407@zju.edu.cn

ORCID iD: 0000-0002-9740-5834

\begin{figure}[H]%
\centering
\includegraphics[width=0.15\textwidth]{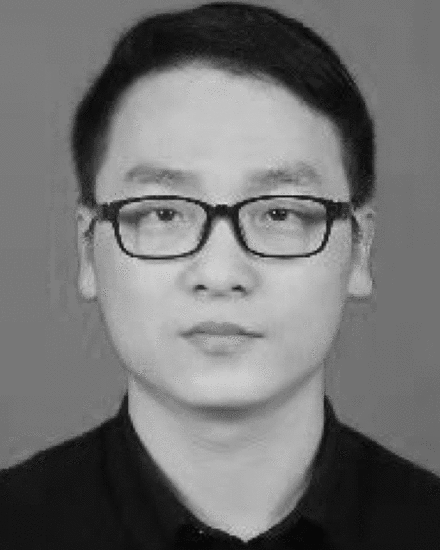}
\end{figure}

\noindent{\bf Haitao Chang }\quad received the B.S., M.S., and Ph.D. degrees in navigation, guidance, and control from Northwestern Polytechnical University, Xi'an, China, in 2010, 2013, and 2018, respectively. He is currently an Assistant Research Professor with the School of Astronautics, Northwestern Polytechnical University. 

His research interests include space robot and control, space teleoperation, and space debris removal, etc.

E-mail: htchang@nwpu.edu.cn

ORCID iD: 0000-0003-4222-4400

\begin{figure}[H]%
\centering
\includegraphics[width=0.15\textwidth]{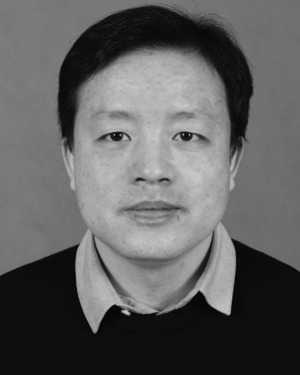}
\end{figure}

\noindent{\bf Zhengxiong Liu }\quad received the Ph.D. degree from Northwestern Polytechnical University, Xi’an, China, in 2012. He is currently an Associate Professor with the School of Astronautics, Northwestern Polytechnical University. 

His research interests include space teleoperation, multibody dynamics, and man–machine interaction.

E-mail: liuzhengxiong@nwpu.edu.cn

ORCID iD: 0000-0002-9427-4066

\begin{figure}[H]%
\centering
\includegraphics[width=0.15\textwidth]{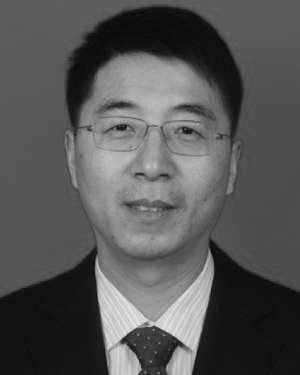}
\end{figure}

\noindent{\bf Panfeng Huang }\quad received
the B.S. degree in test and measurement technology and the M.S. degree in navigation guidance and control from Northwestern Polytechnical University, Xi’an, China, in 1998 and 2001, respectively, and the Ph.D. degree in automation and robotics from The Chinese University of Hong Kong, Hong Kong, in 2005. He is currently a Professor with the School of Astronautics, Northwestern Polytechnical University, and the National Key Laboratory of Aerospace Flight Dynamics, and the Director of the Research Center for Intelligent Robotics, Northwestern Polytechnical University. 

His current research interests include tethered space robotics, intelligent control, machine vision, and space teleoperation.

E-mail: pfhuang@nwpu.edu.cn

ORCID iD: 0000-0002-5132-9602

\end{document}